# Reasoning under Uncertainty: Some Monte Carlo Results

**Paul E. Lehner**
Systems Engineering and C3I Center
George Mason University
Fairfax, VA 22303
plehner@gmuvax2.gmu.edu

**Azar D. Sadigh**
C3I Center
George Mason University
Fairfax, VA 22303

## Abstract

A series of monte carlo studies were performed to compare the behavior of some alternative procedures for reasoning under uncertainty. The behavior of several Bayesian, linear model and default reasoning procedures were examined in the context of increasing levels of calibration error. The most interesting result is that Bayesian procedures tended to output more extreme posterior belief values (posterior beliefs near 0.0 or 1.0) than other techniques, but the linear models were relatively less likely to output strong support for an erroneous conclusion. Also, accounting for the probabilistic dependencies between evidence items was important for both Bayesian and linear updating procedures.

## 1 INTRODUCTION

Reasoning under uncertainty often involves a great deal of judgmental imprecision. The subjective (or data base retrieved) uncertainty estimates that serve as ingredients of an uncertainty calculus are often perceived as arbitrary, imprecise or uncertain. One consequence of this judgmental imprecision is that many decision makers (and researchers) avoid using an explicit uncertainty calculus for fear of being subject to a garbage in-garbage out problem. In this paper this problem is examined by studying the extent to which several different inference procedures are robustly accurate in the context of increasing levels of judgmental imprecision. This paper updates the preliminary report found in Lehner (1990).

## 2 METHOD

A series of monte carlo studies were performed in which we iterated through the following steps.

*Generate "True" Probabilities* - Each run involves an inference problem with one hypothesis (H) node, and either four or seven evidence nodes (A-G). Each node is bi-valued. A "true" probability distribution[1] is defined by randomly assigning (from 0-1 uniform distribution) a value to each instance of $P(H=T)$, $P(A=T|H)$, $P(B=T|HA)$, $P(C=T|HAB)$, etc.[2]

*Assigning Belief Values* - For each probability distribution, a distribution of belief values was generated by adding random error to the true probabilities. For the runs reported in this paper all the error distributions were uniform. Specifically,

$B(x|y) = \text{min} + (\text{max}-\text{min})*\text{RND}$,

where

$\text{min} = \text{maximum}[0.0, P(x|y)-(\text{range}/2)]$
$\text{max} = \text{minimum}[1.0, P(x|y)+(\text{range}/2)]$.

Here $P(x|y)$ is the true conditional probability, $B(x|y)$ is the belief value that is obtained by adding random error to the true probability, RND is a random number and "range" is the range of the error distribution, which we call the error range. When error range = 0.0, the belief values equal the true probabilities. When error range = 2.0, there is no correlation between the true probabilities and the belief values. Our simulation runs used error ranges in the set {0.0, 0.2, 0.4, 0.6, 0.8, 1.0, 1.2, 1.4, 1.6, 1.8, 2.0}.

No matter what the error range, the belief values define a fully-specified and coherent distribution. The inputs to each of the inference procedures were derived from this underlying distribution of belief values.

---

[1] As noted in Lehner (1990) calibration error can be defined without relying the concept of a true probability. Our use of the term "true" is merely a convenience.

[2] A notational note. An expression such $P(A=T|H=T\&BC)$ is a template for a set of expressions. In this case, $P(A=T|H=T,B=T,C=T)$, $P(A=T|H=T,B=T,C=F)$, ... are instances of this template.

*Inference Procedures* - A variety of inference procedures were examined. Each inference procedure prescribes a mechanism for evaluating evidence to generate a posterior belief value. For example, in Bayesian inference the posterior belief in H=T given A=F, B=T, C=T and D=T (i.e., PB(H=T|A=F B=T C=T D=T)) is determined by Bayes rule, whereas the linear models calculate this by a weighted sum of evidence items that support or contradict H=T. The specific inference procedures are described individually in the results section.

*Empirical Evaluation* - For each inference procedure we recorded the distribution of posterior belief values for H=T and H=F. Table 1 gives a typical example using the procedure Proper Bayes. Consider the first cell in this table. It indicates that when error range = 0.0, the expected probability that the posterior belief in H=T is less than .11 when in fact H=T is .068. That is

$$P(PB(H=T)<.11|H=T)=.068.$$

In addition, we also calculated normalized difference between the means (d') between the distributions P(PB(H=T)|H=F) and P(PB(H=T)|H=T). This provides an estimate of the extent to which the inference procedure differentiates between the H=T and H=F condition.

*Analytical Evaluation* - After each monte carlo run we performed an analytical evaluation to explain the results and to identify the extent to which the results are likely to generalize. These analyses were used to design the each new monte carlo run. Rather than execute a series of undirected simulations, each monte carlo run was designed to address a question that the previous analysis left unanswered.

## 3 RESULTS

Presented below are the principal results of this study.

### 3.1 BAYESIAN UPDATING

Bayesianism is a school of thought that argues that a rational system of belief values should conform to the probability calculus and that Bayes rule should be the mechanism by which posterior beliefs are calculated (e.g., Pearl, 1988). In Bayesian theory posterior belief values should be conditioned on all available evidence and all probabilistic interactions among those evidence items should be taken into account. In practice it is often necessary to implement simplified Bayesian procedures that ignore some of the interactions between evidence items. In this study we examined the theoretically correct and two simplified Bayesian procedures.

#### 3.1.1 Proper Bayes

Here we used Bayes rule properly. For instance, in the four evidence case for each evidential state

$$PB(H=T|ABCD) = \sum B(H=T \ \& \ ABCD)/\sum B(HABCD)$$

Table 1 shows the results for three error ranges. Note the shape of the distributions. Proper Bayesian updating consistently resulted in a U-shaped distribution of posterior belief values. This indicates that when H=T, Bayesian updating is more likely to derive an extreme posterior belief value (less than .11) in the wrong direction than an intermediate value (between .44 and .56). The results for H=F are symmetric and are not reported separately.

It is also worth noting that as the error range increased, the proportion of posterior belief values in the middle range (between .33 and .67) did not change substantially. The suggests that proper Bayesian conditioning consistently outputs strong support for some conclusion, and that input error tends to increase the probability that the output values will show strong support for the wrong conclusion.

The U-shaped nature of the posterior belief curve can be accounted for by the fact that Bayesian conditioning is a multiplicative rule applied to likelihood ratios. Whenever the input belief values are very small, there is considerable potential for significant error. We tested this hypothesis by forcing belief values to lie between .05 and .95, and rerunning the simulation. As shown in Table 2, this did have the effect of smoothing out the tail, but as the error range increased the U-shape reappeared.

#### 3.1.2 Naive Bayes

One way to simplify the application of Bayesian inference is to treat the evidence items as though they are conditionally independent of each other. Bayesian models of this type are often called *Naive Bayesian models*. Although recognized as a simplification, this approach does make it easier to apply Bayesian procedures to a large variety of inference problems (e.g., Edwards, et.al., 1968).

Table 1: Average Probability of Posterior Belief Values for Proper Bayes

| Posterior Belief | Four Evidence Items | | | Seven Evidence Items | | |
|---|---|---|---|---|---|---|
| | 0.00 | 0.40 | 1.20 | 0.00 | 0.40 | 1.20 |
| .00-.11 | .068 | .075 | .151 | .047 | .058 | .159 |
| .11-.22 | .050 | .055 | .082 | .031 | .037 | .065 |
| .22-.33 | .048 | .053 | .071 | .029 | .035 | .052 |
| .33-.44 | .049 | .052 | .063 | .030 | .034 | .049 |
| .44-.56 | .053 | .057 | .066 | .033 | .039 | .051 |
| .56-.67 | .063 | .066 | .072 | .039 | .046 | .056 |
| .67-.78 | .080 | .087 | .083 | .054 | .060 | .068 |
| .78-.89 | .122 | .127 | .114 | .088 | .097 | .100 |
| .89-1.0 | .467 | .428 | .298 | .650 | .593 | .399 |
| d' | 1.00 | 0.88 | 0.32 | 1.61 | 1.34 | .046 |

In our study, we examined two forms of naive Bayesian updating. In both cases we used the marginal probability of each evidence item given H=T and H=F to calculate posterior belief. In the case of Simple Naive Bayes, all evidence items were used to calculate posterior belief values. For Strong Naive Bayes posterior belief was calculated in the same way except that for any evidence item x, if $2/3 < B(x|H=T)/B(x|H=F) < 3/2$ then x was dropped from the calculation (i.e., likelihood ratio set to 1).

Table 3 shows the results for some of the Simple Naive Bayes runs. The Strong Naive Bayes results were nearly identical and are not reported separately. In general, the pattern of results were the same as with Proper Bayes, but the differentiation between the two distributions was considerably less. This is reflected in the lower d' values. Surprisingly, when there are seven evidence items, the distribution of belief values remained almost the same. The d' values for Simple Naive Bayes were .68, .60 and .21 which are close to the d' values in Table 3.

Table 2: Average Probability of Posterior Belief Values for Proper Bayes When Extreme Belief Values Are Suppressed

| Posterior Belief | Four Evidence Items | | |
|---|---|---|---|
| | 0.00 | 0.40 | 1.20 |
| .00-.11 | .046 | .053 | .116 |
| .11-.22 | .053 | .058 | .091 |
| .22-.33 | .053 | .057 | .080 |
| .33-.44 | .056 | .059 | .075 |
| .44-.56 | .062 | .065 | .077 |
| .56-.67 | .075 | .077 | .084 |
| .67-.78 | .094 | .098 | .097 |
| .78-.89 | .146 | .145 | .131 |
| .89-1.0 | .414 | .387 | .249 |
| d' | 1.02 | 0.91 | 0.34 |

### 3.2 LINEAR MODELS

In the judgment and decision making (JDM) research literature there is a tradition of research that focuses on the use of linear models to predict human judgment (see Dawes, 1979 for review; Levi, 1989 for recent example). This work is based on two rather surprising but reliable results. First, in probabilistic domains, linear models are often good predictors of human judgment. Even for tasks that appear to be fundamentally pattern-based (e.g., clinical judgment), a regression model of an expert's judgments is usually a good predictor of that expert's future judgments. Second, linear models of expert judgment usually perform better than the experts from which they were derived. This result even holds for linear models with improper equal-weights.

Here we examined three types of linear models.

*Complex Linear.* This technique simply adds up pros and cons and normalizes the result to the number of possible evidence items. A pro was defined as any evidence item for which the conditional probability of that evidence item given the hypothesis and *other evidence items* was greater than one. For instance, for the evidential state A=T, B=F, C=T and D=T, evidence item "C=T" was considered a pro if

$$\frac{B(C=T|H=T, A=T, B=F)}{B(C=T|H=F, A=T, B=F)} > 1.0$$

*Simple Linear.* Same as Complex Liner except that the conditional dependencies between evidence items was ignored. For instance, in the above example "C=T" was counted as a pro if

$$B(C=T|H=T)/B(C=T/H=F) > 1.0.$$

*Strong Linear.* This is exactly the same as Simple Linear except that we did not count an item as a pro or con if the ratio was near 1 (specifically in the range 2/3 to 3/2).

Table 3: Average Probability of Posterior Belief Values for Simple Naive Bayes

| Posterior Belief | Four Evidence Items | | | Seven Evidence Items | | |
|---|---|---|---|---|---|---|
| | 0.00 | 0.40 | 1.20 | 0.00 | 0.40 | 1.20 |
| .00-.11 | .091 | .091 | .131 | .090 | .090 | .138 |
| .11-.22 | .070 | .076 | .095 | .068 | .071 | .094 |
| .22-.33 | .066 | .069 | .088 | .060 | .070 | .085 |
| .33-.44 | .066 | .070 | .085 | .065 | .071 | .081 |
| .44-.56 | .070 | .074 | .086 | .067 | .075 | .082 |
| .56-.67 | .080 | .085 | .089 | .073 | .082 | .085 |
| .67-.78 | .097 | .102 | .101 | .090 | .098 | .099 |
| .78-.89 | .137 | .139 | .122 | .131 | .139 | .123 |
| .89-1.0 | .321 | .294 | .203 | .356 | .304 | .212 |
| d' | 0.60 | 0.54 | 0.19 | 0.68 | 0.60 | 0.21 |

*Weighted Linear.* This was the same as Simple Linear except that a weighting scheme was attached that increased the weight on a pro or con as the likelihood ratios deviated from 1.

Tables 4 and 5 show some of the results for the Complex and Simple Linear procedures. The .000 values in some of the cells is an artifact of the limited number of evidence items, so some posterior belief values are logically impossible. The pattern of results for the Strong Linear and Weighted Linear procedures were similar to that of Simple Linear and are not reported separately. Note that the distributions of posterior beliefs are now single peaked and that strong support for either conclusion is rarely observed. This conservatism is even more apparent when the number of evidence items increases. This reflects the inherent conservatism of an additive versus a multiplicative inference procedure.

On the other hand, when the linear procedure does output strong support for some conclusion it is generally more discriminating than the Bayesian procedures. This is shown in Table 6 which shows the likelihood ratio of posterior belief give H=T vs. H=F. Except for the middle range (.44-.56) where the ratio should be 1.0, the ratio for strong linear (indeed all the linear models) was generally more extreme than that of the Bayesian models. When error range = 0.0, the strong linear model was consistently more extreme. This was often true even when we compared the linear models that ignored evidential interactions to Proper Bayes took account of the conditional dependencies between evidence items.

The d' values reflect the same pattern as the Bayesian models. If conditional dependencies between evidence are taken into account, then d' improves substantially as the number of evidence items increases. If conditional dependencies are ignored, then d' remains nearly the same.

TABLE 4 : Average Probability of Posterior Belief Values for Complex Linear

| Posterior Belief | Four Evidence Items | | | Seven Evidence Items | | |
|---|---|---|---|---|---|---|
| | 0.00 | 0.40 | 1.20 | 0.00 | 0.40 | 1.20 |
| .00-.11 | .006 | .008 | .020 | .000 | .000 | .002 |
| .11-.22 | .054 | .062 | .111 | .004 | .004 | .016 |
| .22-.33 | .000 | .000 | .000 | .023 | .026 | .067 |
| .33-.44 | .200 | .210 | .274 | .083 | .093 | .165 |
| .44-.56 | .000 | .000 | .000 | .192 | .207 | .255 |
| .56-.67 | .347 | .343 | .337 | .283 | .283 | .262 |
| .67-.78 | .000 | .000 | .000 | .253 | .241 | .165 |
| .78-.89 | .295 | .287 | .209 | .134 | .120 | .058 |
| .89-1.0 | .101 | .091 | .050 | .028 | .025 | .010 |
| d' | 0.89 | 0.82 | 0.32 | 1.23 | 1.00 | 0.45 |

TABLE 5: Average Probability of Posterior Belief Values for Simple Linear

| Posterior Belief | Four Evidence Items | | | Seven Evidence Items | | |
|---|---|---|---|---|---|---|
| | 0.00 | 0.40 | 1.20 | 0.00 | 0.40 | 1.20 |
| .00-.11 | .013 | .015 | .025 | .001 | .002 | .003 |
| .11-.22 | .084 | .090 | .131 | .014 | .015 | .025 |
| .22-.33 | .000 | .000 | .000 | .057 | .061 | .092 |
| .33-.44 | .229 | .236 | .286 | .143 | .151 | .197 |
| .44-.56 | .000 | .000 | .000 | .232 | .238 | .263 |
| .56-.67 | .324 | .327 | .325 | .254 | .253 | .236 |
| .67-.78 | .000 | .000 | .000 | .187 | .183 | .131 |
| .78-.89 | .263 | .250 | .188 | .089 | .080 | .045 |
| .89-1.0 | .088 | .082 | .045 | .023 | .017 | .008 |
| d' | 0.63 | 0.56 | 0.19 | 0.66 | 0.59 | 0.19 |

TABLE 6: P(PB(H=T)|H=T)/P(PB(H=T)|H=F) Four Evidence Nodes

| Posterior Belief | Error Range = 0.0 | | | Error Range = 1.2 | | |
|---|---|---|---|---|---|---|
| | Proper Bayes | Naive Bayes | Strong Linear | Proper Bayes | Naive Bayes | Strong Linear |
| 0.0-.11 | 0.145 | 0.285 | 0.103 | 0.517 | 0.654 | 0.513 |
| .11-.22 | 0.420 | 0.521 | 0.178 | 0.719 | 0.796 | 0.558 |
| .22-.33 | 0.596 | 0.675 | 0.311 | 0.830 | 0.844 | 0.689 |
| .33-.44 | 0.777 | 0.796 | 0.567 | 0.882 | 0.933 | 0.851 |
| .44-.56 | 0.972 | 0.980 | 0.985 | 0.976 | 1.005 | 1.001 |
| .56-.67 | 1.280 | 1.204 | 1.783 | 1.099 | 1.064 | 1.202 |
| .67-.78 | 1.650 | 1.452 | 3.119 | 1.208 | 1.194 | 1.391 |
| .78-.89 | 2.367 | 1.920 | 5.510 | 1.378 | 1.280 | 1.761 |
| .89-1.0 | 7.129 | 3.588 | 9.953 | 1.949 | 1.498 | 1.942 |

## 3.3 DEFAULT MODELS

In the AI community there are a number of researchers that advocate the use of qualitative/symbolic approaches to reasoning under uncertainty. The symbolic approach generally involves some form of default reasoning, where a system will jump to default conclusion that can be retracted later if contradictory evidence surfaces (Reiter, 1988).

In our study we examined several default models, all of which involved assigning PB(H=T)=1.0 if any of the LRs exceeded a threshold T and no LRs were less than 1/T; PB(H=T)=0.0 if the reverse occurs; and PB(H=T)=0.5 otherwise. In short, it takes only one piece of confirmatory evidence to jump to a strong conclusion (posterior belief = 1.0 or 0.0), but contradictory evidence leaves the system uncertain (posterior belief = 0.5).

Table 7 shows some of the results for default models where T=3/2 and T=5/2. The principal thing to notice here are the likelihood ratios of posterior belief. When T=3/2, they compare favorably with those generated by the Bayesian procedures, particularly the Naive Bayesian models. In fact the default procedure, using a very mild threshold, was about as discriminating as when the naive Bayes outputs strong (>.89 or <.11) posterior belief values.

Table 7: Average Probability of Posterior Belief Values for Default Models

| Posterior Belief | Threshold = 3/2 | | | Threshold = 5/2 | | |
|---|---|---|---|---|---|---|
| | 0.00 | 0.40 | 1.20 | 0.00 | 0.40 | 1.20 |
| 0.0 | .098 | .110 | .165 | .182 | .207 | .255 |
| 0.5 | .575 | .569 | .597 | .391 | .376 | .411 |
| 1.0 | .327 | .322 | .237 | .426 | .417 | .334 |

Perhaps more surprising is the fact that when the threshold for jumping to a conclusion becomes more stringent (increases to T= 5/2) the default reasoning system was more likely to jump to a default conclusion, but was less likely to be correct when doing so. This counter intuitive result is accounted for by the fact that the thresholds are symmetric. A stringent threshold implies that once a default conclusion has been made, it will take a strong piece of evidence to force a retraction of that conclusion. With a less stringent threshold a weaker evidence item is sufficient to get the system to retract a default conclusion. Consequently, with a less stringent threshold, the default reasoning system is less likely to stick to a default conclusion in the context of contradictory evidence.

## 4 DISCUSSION

### 4.1 ON SCORING RULES

Before discussing the implications of the above results, we need first to address two possible objections to our study. First, why did we not use a proper scoring rule such as mean squared error (MSE) to evaluate the inference procedures? In fact, in our preliminary studies (Lehner, 1990) we did. However as discussed there the expected MSE score for the Bayesian procedures began increased rapidly as the error range increased. When the error range exceeded 1.0, the expected MSE score for all the Bayesian procedures was greater than .25, which is an MSE score that can be guaranteed by ignoring all evidence and asserting PB(H)=.5 under all conditions. In short, the proper scoring rule indicated that in the context of significant higher order uncertainty, Bayesian updating was worse than useless.

This result can now be explained by the U-shaped PB distribution characteristic of Bayesian inference. Other things being equal, squared error rules will tend to attribute higher error scores to techniques that tend toward extreme PB values. Consequently, despite the fact that MSE is a nondistorting scoring rule, it provided a misleading evaluation.

The second objection has to do with the apparent lack of calibration of our Proper Bayes result. A careful reader might proceed through the following line of thought. Consider the error range = 0.0 column in Table 1. If indeed correct Bayesian procedures are being applied to true probabilities, then the posterior belief values should be the true posterior probabilities. This in turn suggests that if $PB(H=T) > .89$, then

$$\frac{P(PB(H=T)>.89|H=T)}{P(PB(H=T)>.89|H=F)} > \frac{.89}{.11} = 8.1$$

Yet Table 1 shows a ratio of only 7.129. Proper Bayes is not calibrated, and therefore incorrectly implemented.

It turns out that the above analysis is wrong, but for a subtle and interesting reason. Since error range is 0.0, we know that within each run $PB(H=T|E) = P(H=T|E)$. Therefore within each run

$$P(PB(H=T)=x \ \& \ H=T)$$

$$= \sum_E P(PB(H=T)=x \ |H=T \ \& \ E) * P(H=T|E) * P(E)$$

$$= \sum x * P(E)$$

here $E|P(H=T|E)=x$

From this we can derive

$$\frac{P(PB(H=T)=x|H=T)}{P(PB(H=T)=x|H=F)} = \frac{E[\Sigma(x*P(E))]/P(H=T)}{E[\Sigma((1-x)*P(E))]/P(H=F)},$$

where $E$ is the expectation *across runs* and $\Sigma$ sums over evidential states (E) *within a run* where $P(H=T|E)=x$.

Now, the values x and P(H=T) are not conditionally independent. Across runs, as P(H=T) is high, the probability of observing an evidential state where $P(H=T|E)>.89$ decreases. Consequently, one should generally expect that

$$\frac{P(PB(H=T)=x|H=T)}{P(PB(H=T)=x|H=F)} < \frac{x*E[P(H=F)]}{(1-x)*E[P(H=T)]}.$$

In our monte carlo runs, $E[P(H=T)]=.5$, so the observed data is consistent with this inequality.

## 4.2 IMPLICATIONS

Some of the implications of this study are listed below.

*Accounting for evidential interactions is essential* - Our simulation results support the common belief that accounting for interactions between the evidence items is essential for effective updating. If interactions are ignored then increasing the amount of evidence available did not substantially improve the ability of either Bayesian or linear updating procedures to discriminate H=T vs. H=F.

*If higher order uncertainty exists, update cautiously* - Bayesian updating tends toward extreme belief values. If higher order uncertainty exists, then the probability that the posterior values will support an incorrect hypothesis increases rapidly.

*Linear updating is accurate, but not very powerful* - The results here suggest that simplistic linear update procedures (e.g., add up pros and cons), are a reasonably accurate approach to reasoning under uncertainty. Relative to Bayesian procedures, linear procedures rarely show strong support for the wrong hypothesis. There are however two provisos. First, interactions between the evidence items must be accounted for. Second, the linear procedures often do not show strong support for *any* hypothesis. When this occurs the reasoner should be prepared to execute a more powerful technique.

*Default reasoning requires asymmetric thresholds* - Implicit in any default reasoning procedure is a conditional probability statement. A knowledge engineer would not add a default rule (if a conclude by-default b) unless the engineer believed that for the set of applications of the system P(b|a) was high. Our results suggest that as the engineer becomes more conservative in defining defaults (i.e., P(b|a)>T and T increases) the default reasoning system may become *more* likely to make erroneous default conclusions. This is because, once the system has made a default conclusion, the conservatism works in reverse. The system is less likely to retract that conclusion. This suggest that default reasoning systems should have asymetric thresholds. Default reasoning procedures that jump to tentative conclusions should be conservative. Procedures that retract default conclusions should be less conservative.

Perhaps the most interesting question that these results suggest is that of the sufficiency of Bayes rule. Consider the following case. A probability model has been specified that we know reflects true probabilities. Upon observing some evidence, E, we apply Proper Bayes and deduce PB(H=T)=.9. However, using a Strong Linear procedure we deduce PB(H=T)=.1. If we accept PB(H=T)=.9, then we are ignoring the output of the inference procedure which we know is, in general, more discriminating. On the other hand, to accept a posterior value other than .9, is to accept a value which we know to be other than the true probability. Furthermore, it suggests that there is more information in the evidence than coherent updating has extracted. We are currently investigating this issue.